\documentclass{article}

\usepackage{PRIMEarxiv}

\usepackage[utf8]{inputenc} 
\usepackage[T1]{fontenc}    
\usepackage[colorlinks=true, linkcolor=black, citecolor=black, urlcolor=black]{hyperref}
\usepackage{url}            
\usepackage{booktabs}       
\usepackage{amsfonts}       
\usepackage{nicefrac}       
\usepackage{microtype}      
\usepackage{lipsum}
\usepackage{fancyhdr}       
\usepackage{graphicx}       
\graphicspath{{media/}}     

\title{

KIRETT: Smart Integration of Vital Signs Data for Intelligent Decision Support in Rescue Scenarios

\thanks{\textit{2024 IEEE International Conference on Electro Information Technology (eIT) | 979-8-3503-3064-9/24/\$31.00 ©2024 IEEE | DOI: 10.1109/eIT60633.2024.10609901
}}}

\author{
  Mubaris Nadeem, Johannes Zenkert, Christian Weber, Lisa Bender, Madjid Fathi \\
  Institute for Knowledge-Based Systems and Knowledge Management \\
  University of Siegen\\
  Hoelderlinstrasse 3, 57068 Siegen, Germany\\
  \texttt{\{Mubaris Nadeem\}mubaris.nadeem@uni-siegen.de} \\
}

\begin{document}
\maketitle

\begin{abstract}

The integration of vital signs in healthcare has witnessed a steady rise, promising health professionals to assist in their daily tasks to improve patient treatment. In life-threatening situations, like rescue operations, crucial decisions need to be made in the shortest possible amount of time to ensure that excellent treatment is provided during life-saving measurements. The integration of vital signs in the treatment holds the potential to improve time utilization for rescuers in such critical situations. They furthermore serve to support health professionals during the treatment with useful information and suggestions. To achieve such a goal, the KIRETT project serves to provide treatment recommendations and situation detection, combined on a wrist-worn wearable for rescue operations.

This paper aims to present the significant role of vital signs in the improvement of decision-making during rescue operations and show their impact on health professionals and patients in need. 

\end{abstract}


\section{Introduction}
In the field of medicine, health professionals are under utmost pressure to provide the optimal treatment for the patient in need. To achieve this, medical personnel are required to consider various sources of information (history data, vital signs, external diagnosis) to provide accurate and patient-oriented treatments to people in need. 
In rescue operations, the time-critical factor adds additional pressure for the health professional to understand, observe and provide the right treatment to the patient in the right time. In this scenario, decision making can prove to be a severe challenge, considering additionally the heightened stress of the rescue personnel with potentially a life on the line. Even further, decisions for seemingly similar cases, may need to vary strongly, depending on contextual factors, as well as changes in vital signs.

Context-based treatment recommendations with additional vital signs integration, can allow rescue operators to focus on the patient, while simultaneously getting relevant information provided to them. This helps to maximize the focus on the patient, while still receiving valuable information, even if situations differ from initial observations and past experiences. Vital signs, such as electrocardiogram (ECG), heart rate, and blood pressure can be used to further narrow down possible remedies to specific medical treatment paths. Vital signs, furthermore, provide the possibility to monitor the patient’s health constantly during the given situation and provide real-time information to the rescue personnel to enable them to adjust their decision-making in real-time. This concludes with the necessity to find a solution that enhances the treatment at hand and minimizes the potential long-term damage based on incorrect treatments.

With the KIRETT project, the goal is to provide new ways to contextualize, assist and potentially optimize time-critical treatments with modern technologies, based on knowledge graph technologies, framing, and utilizing artificial intelligence for a mobile, wearable rescue platform, powered by embedded systems. Aiming to support treatments and actions, multiple knowledge sources (e.g., vital signs) are integrated to assist in the decision-making process of the rescue operator. The base of the treatment recommendations, is a knowledge graph, conceptualized for the rescue scenario and populated from manuals and treatment requirements for rescue operations, outlining treatment paths and standard procedures which must be followed by rescuers~\cite{b1}. These manuals explicitly consider vital signs for decision-making with data thresholds and symptoms. Already today, vital data is provided through stationary devices in the ambulance car. However, displaying vital data on the KIRETT wearable device will help to provide quicker ways to gain necessary data on the patient's status, without the need to consult external medical devices continuously. Furthermore, this provides the possibility to consider the measured vital data to change treatment branches in real-time, based on the processed information.

The goal of this paper is to explore and evaluate a strategy on how to integrate vital signs into the treatment, utilizing and extending the KIRETT knowledge graph. This work will consider the following scientific questions:

\begin{enumerate}
    \item How prevalent are vital signs for the treatment of the patients, based on the existing rescue manual?
    \item To what extend can benefits be derived from the currently available vital data integration for rescue operations?
    \item How can strategies be defined and deployed to skip treatment steps, based on considering vital signs?
    \item How can vital signs be visually presented to the users?
\end{enumerate}

\subsection{The KIRETT-Project}
In the context of the KIRETT project, the objective is to develop a wearable device tailored to be wrist-worn by rescue operators. This device integrates medical data aggregation and processing with artificial intelligence solutions, encompassing situation detection and procedural treatment recommendations facilitated by knowledge graphs (KG). The utilization of a KG enables rescue operators to benefit from data-driven assistance, surpassing conventional patient observation through basic data monitoring. Combining these functionalities into a single device allows health professionals to concentrate more effectively on attending to the immediate needs of the patient. The KIRETT project integrates modules for treatment recommendations, situation detection, data middleware, and embedded device utilization, offering a unique approach to support rescue operations. 

\section{Related Work}
The integration of vital signs into a knowledge graph is a multidisciplinary development in the intersection of healthcare technologies, knowledge management, and rescue operations. This section provides a comprehensive understanding of the given literature and illuminates the state of the art in regard to this paper:
Yu et. al present a framework for predicting emotional stages, involving a knowledge graph and vital signs (e.g., heart rate) from the forehead region \cite{yu2020emotion}. The use of 3D-CNN models allowed the authors to extract the information and integrate the results into a knowledge graph, which allowed knowledge reasoning to predict emotions based on facial videos. 

A vital sign monitor system was developed by Shu et al. which records long-term vitals, like heart rate and breathing rate. The recording is uploaded to a cloud platform enabling visuals on a mobile-based system for the user and managers \cite{8110387}.  Shi et al. use knowledge graphs, in combination with machine learning models, and word segmentation models (BERT) to build an automatized question-and-answer model for medical questions \cite{10327124}. 
The aim here was to provide a medical system to non-medical related users, to gain medical advice and drug recommendations. 

In the clinical environment prediction has been widely researched. So, Jain et. al focuses on the incorporation of medical knowledge through knowledge graphs, which were derived from clinical ontologies (e.g. UMLS). Here the integration of vital signs was achieved, with to aim of enhancing performance in e.g. situations with missing data \cite{jain2023knowledge}. 

The integration of vital signs has been researched in various combinations, focusing on aspects of prediction, question-answer systems, and patient monitoring. However, the scientific addition in this paper fuses the vital signs into active treatment paths in life-threatening situations, assisting with decision-making.

\section{Theoretical Background}

This segment provides an overview of the theoretical background across the essential research areas for this work. 

\subsection{Knowledge Graph}
Handling massive amounts of data, knowledge graphs (KG) are an intelligent solution of reformatting it in a comprehensive way \cite{KG_Tommasini}. As knowledge bases, built in a graph-like structure, they store and represent information of real word entities and their relationships with one another \cite{Building_Healthcare_KG}. Having to handle large amounts of unstructured data, KG are often used in combination with machine learning algorithms and nature language processing (NLP) \cite{KG_Tommasini} to structure sizeable datasets. Within the healthcare domain, Knowledge Graphs (KGs) are experiencing growing prominence, offering a structured framework for medical research and treatment methodologies. Their utilization aims to enhance healthcare delivery by validating diagnoses and formulating tailored treatment plans that align with individual requirements \cite{Building_Healthcare_KG}. A Knowledge Graph (KG) is made up of nodes and edges. Nodes are defined as existing entities, such as medication and treatment steps, while edges provide the possibility to interoperate between n-many nodes to provide meaning between them \cite{nadeem2023kirett}. Graph edges can be either directed or undirected \cite{KRGraphs}.  In addition, KGs can be differentiated into acyclic and cyclic graphs \cite{KRGraphs}, allowing loop functionalities for certain algorithms and vice versa. In the nature of completion, KG can be complete or incomplete, depending on the interconnection level between the existing nodes. If one connection between two nodes is missing, a KG cannot be considered complete anymore \cite{nadeem2023kirett}. This means that the more common type is the incomplete graph since acquiring a dataset containing all connections between node, can prove to be very difficult. A knowledge graph is considered fully connected when it disallows any part of the graph to be unconnected to all other areas. All sub-graphs must therefore be connected to the bigger graph in some way for it to be considered fully connected.

\subsection{Vital signs in Healthcare}
Since the increased popularity of wearable devices in the healthcare and fitness market, continuous monitoring of vital signs (ECG, Heart rate, Blood sugar, etc.) enables the user to control and optimize health factors individually. Health professionals can use such monitoring devices to investigate long-term diseases, such as diabetes, to identify abnormalities in the values to find potential medical issues \cite{khan2016monitoring}. It helps physicians track and provide personalized medical treatment, not only through freshly recorded data but also through long-term measurements throughout a dedicated period of time. Long-term observations can be made for chronic diseases, like cardiovascular disease, which can be monitored intensively due to blood sugar level being checked throughout the day by patients and health professionals. Vital signs can be recorded in different ways, such as through photoplethysmography (PPG) in oxygen saturation \cite{shelley2007photoplethysmography, yu2020emotion}, blood glucose measurement through puncturing the finger \cite{nakayama2007painless} and blood pressure measurements through cuff-pressure \cite{ogedegbe2010principles}. 

\section{Methodology}
Currently, vital signs are recorded from rescue operators through integrated medical devices. Depending on the type of vital signs, continuous monitoring devices are used, where the variation of values is visible in real-time. The analysis of the given values is done by the rescue operators to identify abnormalities and to continuously observe changes within the data which helps to narrow the pool of medical treatments. However, medical vital signs can change within a patient's treatment in a short time, which may not be immediately recognized by the rescue operator. In such a situation, vital signs integrated into a knowledge graph can provide additional information catered to individual steps in treatment for improved overview and accessibility. This enables rescuers to fast reactions and takes away the step of having to check multiple medical devices at every step of a patient's treatment. 

To evaluate the need for vital signs integration into an expert-based knowledge graph for rescue operations, the occurrence of vital signs within the emergency needs to be analyzed:

\subsection{Occurrence of vital signs in emergencies}
The knowledge graph for rescue operations used for project KIRETT, was constructed with the expert-defined manual for treatment paths and standard procedures \cite{b1} for rescue operations. In previous research \cite{nadeem2023kirett} we presented that for the knowledge graph construction vital signs have a major impact. However its widely usage and impact within the navigation through the knowledge graph was not considered. This papers original work focuses on the active navigation purpose of the knowledge graph, allowing skips during treatment, based on vital integration. In addition, it presents the implementation of the KIRETT project. Table \ref{tab1:vitals} provides the occurrence of various vitals throughout said manual's treatment steps, which can be integrated through database entries or real-time recordings with medically certified devices. 

\begin{table}[htbp]
\caption{Occurrence of vital signs in emergency situation, based on the rescue operation manual \cite{b1}	}
\begin{center}
\begin{tabular}{|c|c|c|}
\hline
\textbf{Vitals} & \textbf{Amount} & \textbf{Recording through}\\
\hline
Age 						& 42 	& Database Entry\\ 
Systolic blood pressure 		& 34	& Measurements \\ 
Weight 						& 13	& Measurements \\ 
SPO2 						& 13	& Measurements \\ 
Heart frequency				& 13	& Measurements \\ 
Blood glucose 				& 11 	& Measurements\\
Numeric rating scale for pain& 11	& Database Entry\\
Temperature					& 10	& Measurements \\ 
Glasgow coma scale 			& 8		& Database Entry\\
Endtidal co2 (ETCO2)		& 7		& Database Entry\\
Diastolic blood pressure 	& 5		& Measurements \\ 
Transport 					& 5		& Database Entry\\
Cuff pressure 				& 4 	& Database Entry\\
Time passed					& 3		& Database Entry\\
Respiratory rate 			& 2		& Measurements \\ 
Blood pressure difference 	& 2 	& Measurements \\ 
Burn percentage, Fall altitude 			& 1 	& Database Entry\\
High/Low voltage 		& 1		& Database Entry\\
Hypothermia grade, Krupp grade 			& 1		& Database Entry\\
Heart pulse  				& 1		& Measurements \\ 
QSOFA						& 1		& Database Entry\\
\hline
\end{tabular}
\label{tab1:vitals}
\end{center}
\end{table}

Table \ref{tab1:vitals} shows, that vital signs, like age, systolic blood pressure, weight and oxygen saturation are frequently used in the treatment paths and standard working procedures of rescue operations, which indicates that the rescue operator needs to observe the values during a treatment. This means, that vital parameters and database-entry parameters are actively involved in the decision-making of rescue operations. The manual is divided into treatment paths and standard working procedures. Observing these values in the context of the whole treatment manual, it becomes apparent, that in 78 percent (29/37 treatment paths) of treatment paths and 51 percent (20/39 standard procedures) of standard procedures need input from vital signs either from database entries or real-time measurements for various steps they contain. This proves the significant need for vital integration into the treatment paths during rescue operations. 

Further involvement of vital signs in the life cycle of a rescue operation is within the calculation of drug dosage during a rescue operation. Depending on the age or weight different medication need to be provided and prepared for the patient. In the manual \cite{b1} following occurrences of medication preparation is mentioned: SPO2 (mentioned 13x | recording through measurements), Weight (mentioned 11x | recording through Database Entry) and Time passed (2x | recording through measurements). SpO2 for instance can be used to calculate the medication, which needs to be provided to the patient. Such usage of vital signs clearly demonstrates the imperative inclusion of vital signs in medication dosage calculations. Moreover, treatment steps may differ based on the age of the patient, this can for example be encountered in treatment paths like reanimation or seizure. In such scenarios previously recorded vitals can provide navigational support through the treatment. 
The potential for automated computation of these parameters presents a viable solution to mitigate inaccuracies in medication dosing, enabling healthcare professionals to direct their attention toward patient needs and expedite treatment delivery.

\section{Implementation}
The following section will go into the implementation of the vital signs being integrated into the KIRETT project, explaining of how this interaction between various modules within the KIRETT wearable works: 

\subsection{Visual integration of Vital signs in treatment graph}
\begin{figure}[htbp]
\centering
\includegraphics[width=220px]{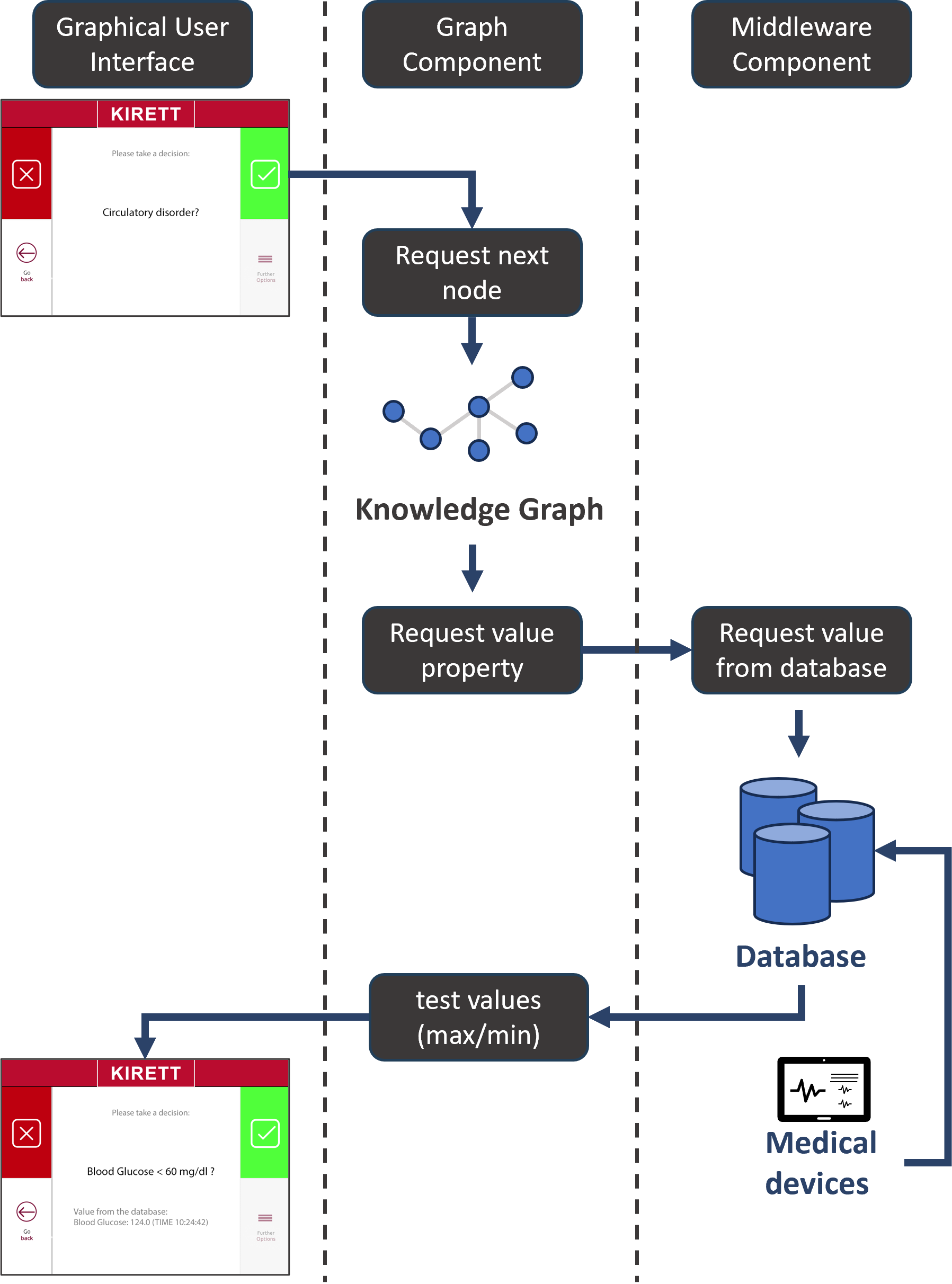}
\caption{This figure outlines the integration of vital signs into a knowledge graph through three modules: GUI, Graph, and Middleware. The GUI visually displays treatment steps with vital signs, sourced from the database. The Graph hosts the knowledge graph, providing content for visualization and managing data retrieval from the database. The Middleware component stores values from medical devices and control center information in a structured database, facilitating data retrieval for inquiries through the graph.}
\label{workflow}
\end{figure}

While the patient monitor of the KIRETT wearable already provides a good overview on the patient and their vitals, sometimes it may additionally prove useful to display them to a rescuer during the treatment in areas of the knowledge graph where it may be relevant.
An example of such a node where it may be beneficial, would be a step-in treatment that contains a decision that depends on certain vital signs or information of the patient. This may apply to nodes holding treatments or procedures based on certain vital signs such as blood sugar or SpO2, or decisions in treatment that need to be made by the rescuer based on a value.
Another application where retrieval and display of vitals could be beneficial, would for example be to determine which medication dosage needs to be given to a patient depending on their weight and age.
To make accurate decisions, rescuers will need this specific value at this point in time to appeal to patient needs~\cite{nadeem2023kirett}. This may be difficult to do with values having to be read from various sources to gain a full picture. So instead of having to consult external devices, the current value will automatically be retrieved from these devices by the Middleware via \textit{Bluetooth} to simplify the process for rescuers~\cite{nadeem2023kirett}.

To be able to display this value at the step where it is required, a few of the modules of project KIRETT have to work together closely~\cite{nadeem2023kirett}. The resulting workflow as can be seen in figure~\ref{workflow}, will be elaborated on in the following:
The interaction of the modules begins in the graphical user interface (GUI), where the current step in treatment is displayed to the user. The rescuer presses a button to proceed to the next node (Fig. \ref{workflow}, first level), which will be the step requesting the vital signs in this example. The GUI sends a message requesting the next node to the Graph component, which queries its KG of treatment steps built from the operation manual for rescue services 2020~\cite{b1}.
When the \textit{Cypher} query returns from the \textit{Neo4J} graph database, it delivers all properties attached to the current node to the Graph component (Fig. \ref{workflow}, third level).
This process happens in every step where the next node is requested, which needs to be displayed by the GUI again. When patient vitals are involved, there will be the value property among all the others, which lets the Graph know that it needs to make an additional call that is required for the rescuer to make an accurate decision on what needs to be done next. 

The Graph notices that in this node, a property is required for the rescuer to make an accurate decision on what needs to be done next. It therefore sends a message to the Middleware component, which hosts the database.
When the value is requested there, it will be retrieved from medically certified devices such as the \textit{ZOLL X-Series} via \textit{Bluetooth} connection.

\begin{figure}[htbp]
\includegraphics[width=180px]{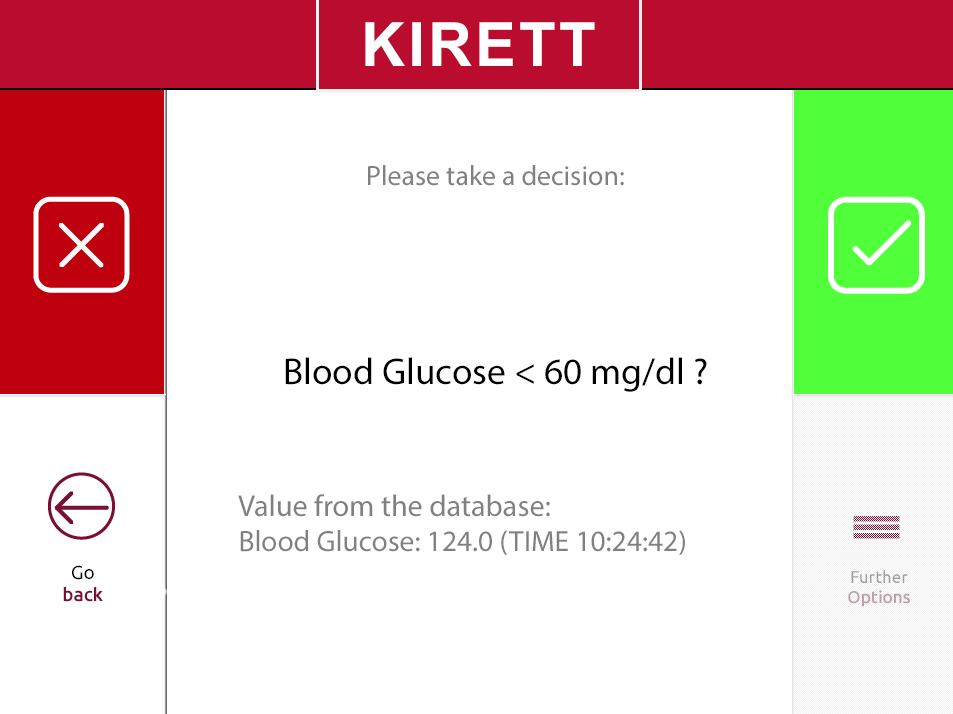}
\centering
\caption{This figure presents the visualization of vital signs on the KIRETT wearable. Providing buttons to accept and decline, the Graphical User Interface (GUI) provides the ability to actively decide whether the value provided in the treatment step matches with the data from the database. This is an additional security layer for the rescue operator, to actively change, observe and decide on the best treatment of the patient.}
\label{bloodsugar}
\end{figure}

Once the Middleware has acquired the value, it will return the results to the Graph component, where they are ordered and packed up with all the other variables needed to display the node on the GUI. This returning message from the Graph will then trigger the GUI to display the new node, this time alongside the vital signs it has received (Fig. \ref{bloodsugar}).

One such node may request multiple values at once, should it contain more than one requirement in its treatment step.
All values this node may have requested, are listed beneath each other on the GUI, and additionally carry a timestamp for added transparency on when this value was fetched from the database. This provides the rescue operators with insight on whether this value is very recent and therefore trustworthy, or whether they should make sure by consulting external devices.
Should a value not be available at a certain time, the word "unknown" will be displayed in its stead to give the rescuers a hint that this value may be required and is not available to the device at the time.

This is the workflow of the communication between the components to retrieve and display patient vital signs to the user where needed.

\subsection{Skip treatment steps, based on vital data-integration}
This concept, however, can easily be expanded to provide the rescuers with even more help by adding an indicator to the displayed value, such as a certain text color that lets the user know whether the value is within the prescribed parameters, or lies outside safe values.
This can be done by the Graph component before it sends back the values it acquired from the Middleware to the GUI. Once it has obtained the patient vital signs from the Middleware, it can consult the corresponding node of the knowledge graph for the span its values should be in. These indicators are described as min and max.
They can be utilized to make use of the newly acquired value to already make an assessment whether the result lies within the parameters of the question or may be a reason for concern. This information, alongside all other data needed to display the next node, would then get sent back to the user interface to be displayed there. Very concerning values may even be a reason to cause a warning to pop up to the user to ensure they notice the critical state of their patient.
And, to take this idea even one step further, the Graph could use this gathered information, and consideration on whether this value fits the one prescribed in the treatment step, to make an automated decision if a value indicates a very clear direction for the pending decision. Such a decision would need to have clear transparency for the user that it has taken place, to allow them to undo this automated conclusion, should they desire to do so. If this is applicable, such automated decision making within the Graph component may benefit users greatly in speeding up decision processes, and therefore allowing faster treatment of patients, and quick reaction to sudden changes. To improve this process of automated support in decision-making, data can be re-utilized in future treatments, to calculate probabilities for future knowledge fusion (e.g. with the help of bayesian networks). Additionally, to this advantage using gathered information, sharing said data also allows a better follow-up treatment in the hospital and general practitioners' offices.

\section{Discussion}
The integration of vital signs data into knowledge graphs for intelligent decision support in rescue operations represents a significant advancement in the field of emergency medical care. In the study of the paper, vital signs data from Bluetooth-connected devices were integrated in a wearable device utilizing a treatment graph for rescue operations as shown in Figure~\ref{workflow}. By incorporating vital signs data into a knowledge graph structure, medical rescue staff, doctors and experts can access relevant and timely information to inform their decision-making process during critical situations. In our evaluation we focused the two research questions a) How does the integration of vital signs data into knowledge graphs improve the efficiency and effectiveness of decision-making in rescue operations? and b) What are the technical challenges associated with the integration of Bluetooth-connected devices with the KIRETT wearable technology for vital signs monitoring in dynamic environments?

Following these two main research questions, in our evaluation, we observed several advantages regarding the knowledge-graph-supported vital sign data integration:
\begin{itemize}
    \item Incorporating vital signs into a knowledge graph gives rescue teams a complete picture of the patient's condition, improving situational awareness for prompt and informed decision-making, resulting in more effective treatments.
    \item With real-time vital signs data from the KIRETT wearable device, rescue operations can optimize resource allocation, particularly in mass casualty incidents. Prioritizing patients based on condition severity ensures immediate attention for critical cases.
    \item Incorporating vital signs data into the treatment graph improves communication and collaboration among medical staff. The integrated data can be easily shared between the rescue team and hospital, ensuring seamless follow-up treatment, with optimal provision of data records from the rescue phase.
    \item Utilizing knowledge graphs to integrate vital signs data empowers data-driven decision-making in rescue operations. Analysis of patterns and trends within the data enables medical staff to identify potential risks and anticipate complications, facilitating proactive interventions to mitigate adverse outcomes.
    \item The structure build with nodes and edges of knowledge graphs enables the system to adapt to evolving needs and scale according to the size and complexity of rescue operations. As new devices and technologies emerge, the framework can easily incorporate additional data sources, ensuring that the medical staff has access to the most comprehensive information available.
\end{itemize}

Considering the first question, the effectiveness and efficiency of the whole medical treatment, from the first response to the handover in the hospital, vital signs have a great impact of the rescue service. Especially in the field of assistive decision making, treatment recommendations, based on vital signs can support rescue personnel in their daily work. For future work, the use of Bluetooth devices, needs further evaluation. In dynamic environments, such devices may have communication issues. In summary the integration of vital signs data into knowledge graphs holds immense potential for improving decision support in rescue operations.

\section{Conclusion}
The integration of vital sign parameters into knowledge graphs, as investigated in the KIRETT project, represents an approach to improving decision support in rescue operations. The results present that there is a significant need for the integration of vital signs into treatment pathways, which has a considerable impact on the decision-making process. The visual integration of vital signs into the treatment path not only provides a comprehensive overview of the patient's condition but also enables medical staff to make informed decisions quickly and efficiently in critical situations.

The benefits of this integration extend to improved situational awareness, resource allocation, communication, and data-driven decision-making. The ability to skip treatment steps based on real-time vital data further increases efficiency and can save crucial time in life-saving scenarios. The workflow presented, involving the graphical user interface (GUI), graph component, and Middleware, demonstrates a seamless process for retrieving and displaying vital signs data where and when it is needed.

\section{Future Work}
\begin{figure}[htbp]
\includegraphics[width=190px]{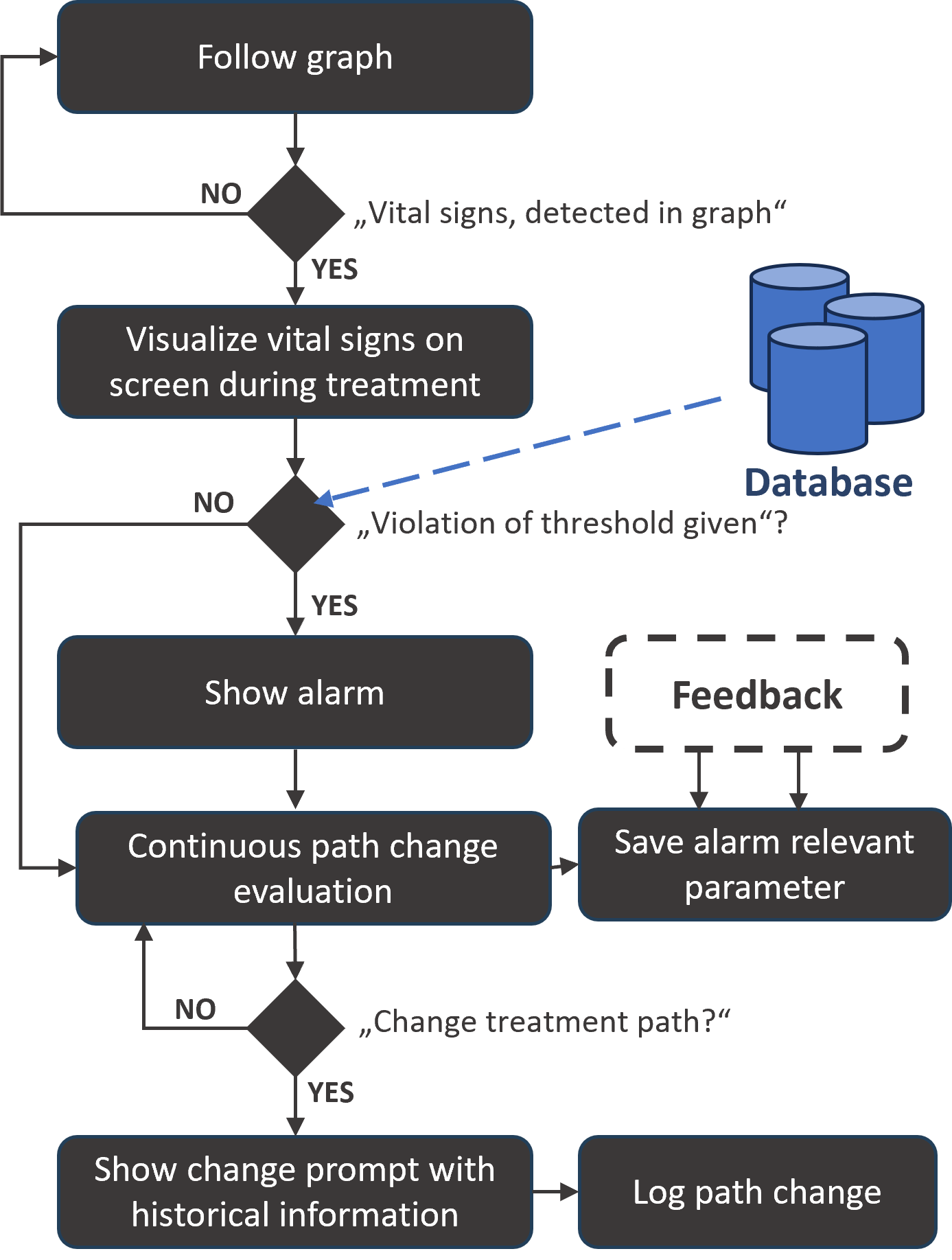}
\centering
\caption{This figure outlines a proactive notification system designed to adjust treatment paths based on historical data and vital signs. When a threshold is breached, an alarm is activated, storing the current vital signs and treatment step. Rescue operators can review the alarm and decide whether to modify their treatment path approach accordingly.}
\label{ausblick}
\end{figure}

The integration of vital signs into a knowledge graph for intelligent decision support in rescue operations is a promising path for further research and development. In the future, it can be investigated how machine learning algorithms analyze treatment patterns and trends with the help of vital signs. The prediction models, using e.g. Bayesian network can support decision-making for rescue operations. An alarm notification system for treatment change assistance (Fig. \ref{ausblick}) can be developed, in which a threshold violation could lead to a visualization of the alarm, providing the increased vital signs with its threshold from the rescue operator's manual at hand. Based on health professionals feedback, values can be stored and used for an evaluation of the correctness of the current path change. If a treatment path modification is needed, health professionals could decide and get a visualization of the changed treatment path with additional, reasoned, historical data for clarification.  With such an adjustment, rescue operation assistance tools, like KIRETT, provide a continuous reevaluation of the stored, freshly recorded and threshold providing data to enable an explainability for possible treatment changes.

In addition, the interoperability between multiple systems can be investigated, focusing on standardized formats and integration of multiple knowledge sources, like Electronic Health Records (EHR) or electronic patient records (EPR). A usability study of the vital integration can give a closer insight into the practical implementation of the system in real-world scenarios, which would result in iterative improvements.

In the long term, data can be recorded, anonymized, and then used to analyze the integration factor of the vitals, allowing the identification of trends and correlations, which may not be directly seen by health professionals.

\section*{Funding}
Ongoing research is financially supported by the German Federal Ministry of Education and Research (BMBF) and coordinated by CRS Medical GmbH (Aßlar, Germany) and mbeder GmbH (Siegen, Germany).

\bibliographystyle{unsrt}  
\bibliography{templateArXiv}

\end{document}